\IfFileExists{ieeeconf.cls}{%
  \documentclass[letterpaper,10pt,conference]{ieeeconf}%
}{%
  \documentclass[letterpaper,10pt,conference]{IEEEtran}%
  \providecommand{\overrideIEEEmargins}{}%
}

\IEEEoverridecommandlockouts
\usepackage{amsmath,amssymb}
\usepackage{graphicx}
\usepackage{booktabs}
\usepackage{multirow}
\usepackage{array}
\usepackage{makecell}
\usepackage{xcolor}
\usepackage{cite}
\usepackage{url}
\usepackage{placeins}
\usepackage{stfloats}
\usepackage{microtype}

\newcommand{\TNVDConfig}{\ensuremath{B=24,\ H=18\,\mathrm{m},\ N_R=20,\ N_S=60,\ K=25}}

\newcommand{\NCLTSCMean}{73.73}

\newcommand{\NCLTTNVDMean}{85.14}
\newcommand{\NCLTGain}{11.41}


\newcommand{\ours}{TNVD-SC}

\newcommand{\JS}{Jensen--Shannon}

\newcommand{\figorplaceholder}[3]{%
  \IfFileExists{#1.pdf}{%
    \includegraphics[width=#2]{#1.pdf}%
  }{%
    \IfFileExists{#1.png}{%
      \includegraphics[width=#2]{#1.png}%
    }{%
      \IfFileExists{#1.jpg}{%
        \includegraphics[width=#2]{#1.jpg}%
      }{%
        \fbox{\parbox[c][#3][c]{#2}{%
          \centering\small Figure placeholder\\[2pt]%
          \texttt{\detokenize{#1.(pdf|png|jpg)}}%
        }}%
      }%
    }%
  }%
}

\title{TNVD-SC: Diagnosing Scan Context Failures in Forests with Terrain-Normalized Vertical Distributions}

\author{Jineon Kim$^{1}$ and Jhonghyun An$^{1,*}$%
\thanks{This work has been submitted to the IEEE for possible publication. Copyright may be transferred without notice, after which this version may no longer be accessible.}%
\thanks{$^{1}$School of Computing, Gachon University, Seongnam 13120, Republic of Korea.}%
\thanks{$^{*}$Corresponding author ({\tt\small jhonghyun@gachon.ac.kr}).}%
}

\begin{document}
\maketitle
\thispagestyle{empty}
\pagestyle{empty}

\begin{abstract}
Long-term LiDAR place recognition in forests is difficult because repetitive vegetation and uneven terrain create severe structural aliasing. We diagnose Scan Context (SC) by separating candidate retrieval from fine matching: on the standardized TreeLoc++ Wild-Places protocol, a TNVD support key places a valid location in the top-25 set for approximately 99\% of queries, while conventional SC matching on those candidates reaches only 61.84\% mean R@1. With an identical raw-SC candidate set, terrain normalization improves max-height cells by only 1.21 points but vertical-distribution cells by 19.72 points. These observations motivate Terrain-Normalized Vertical Distribution Scan Context (\ours{}), which preserves the SC ring--sector topology while encoding terrain-relative vertical probability distributions. Under the standardized TreeLoc++ protocol, \ours{} reaches 91.83\% mean R@1 and attains the highest R@1 mean and R@1 stability among the compared methods, while TreeLoc++ remains stronger in mean MR, MF1, and AUC. The unchanged method reaches 85.14\% mean R@1 on the four-session CVTNet-style NCLT evaluation. A lossless count-coded implementation with exact hierarchical Jensen--Shannon search reduces descriptor payload by 3.86$\times$ and compute latency by 23.8\% without changing the audited retrieval outputs.
\end{abstract}

\begin{figure*}[!t]
\centering
\includegraphics[width=0.87\textwidth]{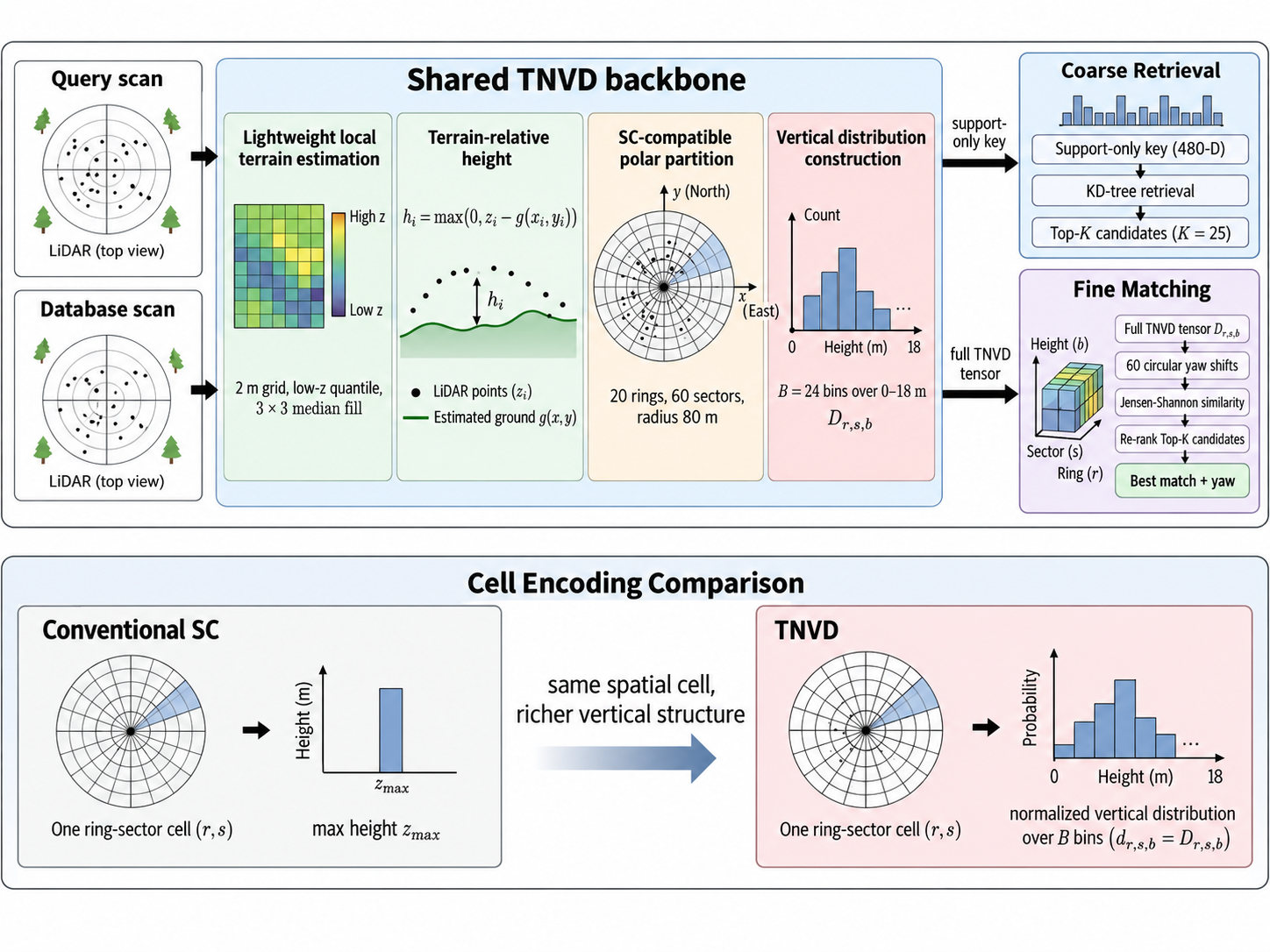}
\caption{Overview of \ours{}. Query and database scans share the TNVD backbone. A 480-D support-only key retrieves top-$K$ candidates, while the ordered full TNVD tensor performs circular fine matching. Bottom: conventional SC stores one maximum height per cell, whereas TNVD retains its terrain-relative vertical distribution.}
\label{fig:architecture}
\end{figure*}

\section{Introduction}
Reliable place recognition is a core component of long-term robotic autonomy, supporting loop closure, global relocalization, and recovery after localization failure. Forests are particularly difficult: tree stems repeat at similar spacings, foliage and understory vary over time, visibility changes with viewpoint and season, and terrain relief changes the apparent vertical arrangement of otherwise similar places. Wild-Places was introduced specifically to expose these challenges in unstructured natural environments~\cite{knights2023wildplaces}.

Scan Context (SC)~\cite{kim2018scancontext,kim2022scancontextpp} remains attractive for field robots because it is training-free, requires no site-specific retraining, and preserves a simple coarse-to-fine retrieval structure suitable for CPU-class deployment. In our standardized comparison (Table~\ref{tab:wildplaces_main}), every learned baseline has an R@1 stability ratio below 1.85, whereas \ours{} exceeds 2.8 on both sites, making pair-to-pair consistency an additional practical consideration. A LiDAR scan is divided into polar ring--sector cells, each cell is compressed to a maximum-height scalar, a yaw-invariant key retrieves candidates, and circular sector shifts resolve yaw. The compression is severe: all vertical structure inside a cell is reduced to one extreme value. In forests, different places can therefore produce similar SC patterns whenever their upper envelopes align even though their internal vegetation layers differ.

This ambiguity is not removed simply by widening retrieval. Figure~\ref{fig:candidate_budget} shows that, even at $K=250$, SC candidate recall reaches 97.45\% on Venman and 98.47\% on Karawatha while final R@1 remains near 40--50\%. We therefore treat retrieval and fine matching as separate failure modes rather than assuming that a stronger candidate search alone will resolve structural aliasing.

Recent forest place-recognition methods use learned BEV density representations~\cite{shen2025forestlpr}, hierarchical transformers~\cite{griffiths2025hotformerloc,griffiths2026hotflocpp}, explicit tree geometry~\cite{jung2026treelocpp}, and training-free signal-processing approaches~\cite{joseph2025matched}. Handcrafted descriptors have also explored richer height or probabilistic structure~\cite{bae2025csc,xu2022pghci,lee2026probe}. Thus, vertical distributions, terrain normalization, and probability divergences are not claimed as individually new. Our question is diagnostic: \emph{at which stage does SC fail in long-term forests, and what representation change is justified by that failure?}

We use \emph{TNVD} to denote the terrain-normalized vertical-distribution cell representation and \emph{TNVD-SC} to denote the complete place-recognition system consisting of TNVD construction, support-key retrieval, and circular distribution matching. Two controlled observations motivate the system. First, the TNVD support key places a valid location in the top-25 set for 98.89\% of Venman and 98.76\% of Karawatha queries, yet conventional SC matching on those candidates reaches only 57.20\% and 66.47\% R@1. Second, when the candidate set is fixed to the \emph{same raw-SC ring-key Top-25 set}, terrain normalization changes max-height mean R@1 by only +1.21 points but changes the vertical-distribution representation by +19.72 points. The benefit of terrain normalization is therefore strongly conditioned on what information the cell representation preserves.

These findings motivate \ours{}. The method keeps the SC range--azimuth topology and exhaustive circular yaw search, but replaces the per-cell maximum with a terrain-relative vertical probability distribution. A 480-D support-only key is used only for coarse retrieval; the full ordered tensor is retained for fine re-ranking. The frozen method is implemented efficiently through lossless integer count coding and exact hierarchical Jensen--Shannon pruning, preserving all audited retrieval outputs while reducing memory and latency.

Our contributions are:
\begin{itemize}
\setlength{\itemsep}{1pt}\setlength{\parskip}{0pt}\setlength{\parsep}{0pt}
    \item \textbf{Stage-wise failure diagnosis.} We separate SC candidate retrieval from fine discrimination and show that retrieving a valid candidate does not by itself resolve forest aliasing.
    \item \textbf{Controlled representation--terrain interaction.} With identical raw-SC candidates across all conditions, terrain normalization provides a marginal gain for max-height cells but a much larger gain for vertical-distribution cells, revealing an 18.51-point interaction in mean R@1.
    \item \textbf{Diagnosis-driven TNVD-SC.} We derive a training-free SC-compatible system with a 480-D support retrieval key and circular distribution matching, reaching 91.83\% mean R@1 on standardized Wild-Places and 85.14\% on the four-session CVTNet-style NCLT evaluation.
    \item \textbf{Exact efficient implementation.} Lossless count coding and exact hierarchical Jensen--Shannon search reduce persistent descriptor payload by 3.86$\times$ and compute latency by 23.8\% without changing the audited retrieval decisions.
\end{itemize}

\section{Related Work}
\subsection{Handcrafted and Probabilistic LiDAR Descriptors}
SC~\cite{kim2018scancontext} introduced polar max-height cells with yaw-invariant retrieval and circular matching, while SC++~\cite{kim2022scancontextpp}, RING++~\cite{xu2023ringpp}, and SOLiD~\cite{kim2024solid} pursue complementary invariances or compact encodings. Matched Filtering~\cite{joseph2025matched} uses a training-free BEV representation with two-stage correlation for place recognition and relative pose; its Wild-Places numbers use the original 3\,m protocol and are therefore not mixed with our 5\,m TreeLoc++ comparison. Richer vertical/probabilistic representations include PGHCI~\cite{xu2022pghci}, CSC~\cite{bae2025csc}, and PROBE~\cite{lee2026probe}. These works motivate richer structure, but our focus is the controlled diagnosis of which SC stage fails and which SC-compatible representation addresses that failure.

\subsection{Learned and Forest-Specific Recognition}
Learned LiDAR retrieval spans PointNetVLAD, MinkLoc3D/v2, TransLoc3D, LoGG3D-Net, OverlapTransformer, SeqOT/CVTNet, BEVPlace++, and OverlapMamba~\cite{uy2018pointnetvlad,komorowski2021minkloc3d,komorowski2022minkloc3dv2,xu2023transloc3d,vidanapathirana2022logg3d,ma2022overlaptransformer,ma2023seqot,ma2023cvtnet,luo2025bevplacepp,luo2025overlapmamba}. Forest-specific methods include ForestLPR, HOTFormerLoc/HOTFLoc++, and TreeLoc++~\cite{shen2025forestlpr,griffiths2025hotformerloc,griffiths2026hotflocpp,jung2026treelocpp}. Robust terrain estimators such as Patchwork/Patchwork++~\cite{lim2021patchwork,lee2022patchworkpp} further show that local ground modeling is well studied. We therefore do not claim terrain estimation itself as novel; \ours{} uses one frozen lightweight reference to isolate its interaction with the downstream cell representation.

\section{Diagnosing Scan Context in Forests}
\subsection{Retrieval and Fine Matching Are Distinct Bottlenecks}
We decompose an SC-style pipeline into (i) descriptor construction, (ii) yaw-invariant candidate retrieval, and (iii) circular fine matching. A positive that never enters the candidate list and a positive that is retrieved but then loses to an aliased forest location are different failure modes.

\begin{figure}[!t]
\centering
\figorplaceholder{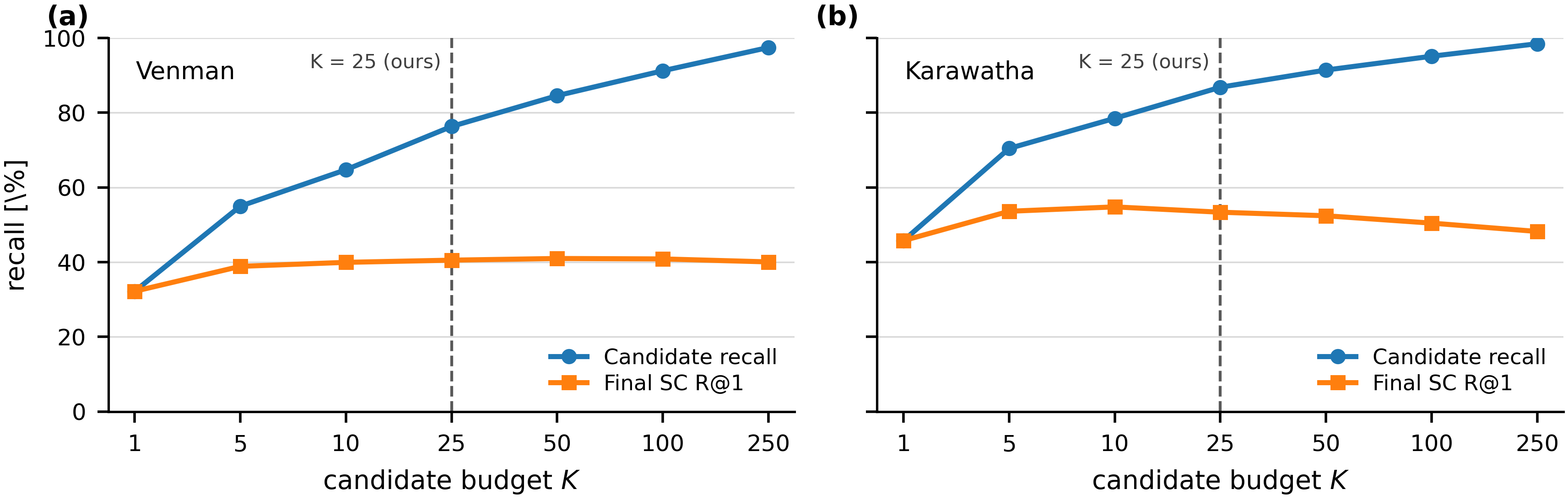}{0.88\columnwidth}{1.18in}
\caption{Candidate-budget diagnosis for conventional SC. Candidate recall continues to rise with $K$, whereas final SC R@1 saturates. The single dashed line marks the $K=25$ operating point used by \ours{}.}
\label{fig:candidate_budget}
\end{figure}

Figure~\ref{fig:candidate_budget} first varies the SC candidate budget. Candidate recall continues to rise as $K$ increases, reaching 97.45\% on Venman and 98.47\% on Karawatha at $K=250$, while final SC R@1 remains near 40--50\%. Increasing retrieval breadth therefore does not remove the fine-matching failure. Table~\ref{tab:component} then controls retrieval and matching directly. The TNVD support key places a valid location in the top-25 set for 98.89\% and 98.76\% of queries, yet the SC matcher on those candidates reaches only 57.20\% and 66.47\% R@1. Conversely, the TNVD matcher improves the fixed SC candidate set to 68.72\% and 79.15\% but remains bounded by its lower candidate recall. Only the full TNVD-SC combination reaches 92.24\% and 91.41\%. Candidate recovery and fine discrimination are therefore distinct, non-substitutable bottlenecks.

\begin{table}[!t]
\centering
\caption{Component decomposition on standardized Wild-Places. R@25 is candidate recall before fine reranking.}
\label{tab:component}
\scriptsize
\setlength{\tabcolsep}{3.2pt}
\begin{tabular}{@{}llccccc@{}}
\toprule
Retrieval & Matcher & \multicolumn{2}{c}{Venman} & \multicolumn{2}{c}{Karawatha} & Mean \\
 & & R@1 & R@25 & R@1 & R@25 & R@1 \\
\midrule
SC ring key & SC & 40.53 & 76.32 & 53.37 & 86.82 & 46.95 \\
SC ring key & TNVD & 68.72 & 76.32 & 79.15 & 86.82 & 73.94 \\
TNVD support & SC & 57.20 & \textbf{98.89} & 66.47 & \textbf{98.76} & 61.84 \\
\textbf{TNVD support} & \textbf{TNVD} & \textbf{92.24} & \textbf{98.89} & \textbf{91.41} & \textbf{98.76} & \textbf{91.83} \\
\bottomrule
\end{tabular}
\end{table}

\subsection{Terrain Normalization Depends on the Cell Representation}
To isolate terrain normalization from candidate retrieval, we generate the raw-SC ring-key Top-25 candidates once and keep that candidate set identical in all four conditions of Table~\ref{tab:terrain_interaction}. Only the cell representation and whether the same local terrain reference is applied are changed. Terrain normalization raises mean R@1 for max-height cells only from 46.95\% to 48.16\% (+1.21 points), whereas the vertical-distribution representation rises from 54.22\% to 73.94\% (+19.72 points). The resulting representation-dependent interaction is +18.51 points.

The difference is consistent with the information retained by each representation. A max-height cell keeps only one upper-envelope scalar, so removing low-frequency terrain relief changes that already compressed cue only modestly. A vertical distribution instead preserves multiple height layers; without normalization, terrain offsets translate probability mass along the height-bin axis and can break bin correspondence across revisits. Terrain-relative coordinates align understory, stem, and canopy layers before distribution matching. Terrain normalization is therefore not an independently sufficient explanation for the gain; its value depends strongly on the representation that follows it. Because all four conditions share the raw-SC candidate set, their achievable R@1 is capped by candidate recall (76.32\% on Venman and 86.82\% on Karawatha). The terrain-normalized vertical-distribution matcher already reaches 90.0\% and 91.2\% of these ceilings, respectively, indicating that the controlled interaction measurement is itself retrieval-ceiling limited.

\begin{table}[!t]
\centering
\caption{Terrain normalization under an identical raw-SC Top-25 candidate set (pair-mean R@1, \%).}
\label{tab:terrain_interaction}
\scriptsize
\setlength{\tabcolsep}{4.0pt}
\begin{tabular}{@{}lcccc@{}}
\toprule
Cell representation & Terrain norm. & Venman & Karawatha & Mean \\
\midrule
Max-height & No  & 40.53 & 53.37 & 46.95 \\
Max-height & Yes & 42.65 & 53.67 & 48.16 \\
Vertical distribution & No  & 50.66 & 57.78 & 54.22 \\
\textbf{Vertical distribution} & \textbf{Yes} & \textbf{68.72} & \textbf{79.15} & \textbf{73.94} \\
\bottomrule
\end{tabular}
\end{table}

\section{Terrain-Normalized Vertical Distribution Scan Context}
Figure~\ref{fig:architecture} summarizes the diagnosis-driven design. We use \emph{TNVD} for the cell-level terrain-normalized vertical-distribution representation and \emph{TNVD-SC} for the complete retrieval-and-matching system. \ours{} retains the original SC range--azimuth partition and exhaustive circular yaw search while changing the vertical coordinate, per-cell encoding, retrieval key, and fine matcher.

\subsection{Lightweight Terrain-Relative Height}
Let a LiDAR scan be $\mathcal{P}=\{\mathbf{p}_i\}_{i=1}^{N}$ with $\mathbf{p}_i=(x_i,y_i,z_i)$. We estimate a lightweight local terrain reference $g(x,y)$ on a Cartesian grid with 2\,m cells. Cells containing at least three in-range points obtain a provisional terrain estimate from the 10th percentile of their $z$ values. Each grid cell is then assigned the median of available provisional estimates within its $3\times3$ neighborhood, simultaneously smoothing valid estimates and filling sparse cells. If no valid provisional estimate exists in that neighborhood, a scan-level 10th-percentile height is used as a fallback. The terrain-relative height is
\begin{equation}
    h_i = \max\left(0,\ z_i-g(x_i,y_i)\right).
    \label{eq:terrain_height}
\end{equation}
The fixed parameters are engineering choices: the 2\,m grid retains local relief while limiting sparsity, the low percentile is more robust to vegetation than a single minimum, the three-point minimum avoids extremely small local samples, and the $3\times3$ median/fill suppresses isolated outliers and fills local holes. More sophisticated LiDAR ground estimators exist~\cite{lim2021patchwork,lee2022patchworkpp}; ground segmentation itself is not our contribution.

An audit confirms that the global fallback has negligible influence on the descriptor-relevant points: it is used for 0.000248\% on Wild-Places, 0.000007\% on Oxford Forest, and 0.060355\% on NCLT. Even on NCLT, the median and 95th-percentile per-scan fallback ratios are only 0.0583\% and 0.1850\%, respectively. The same estimator and parameters are frozen in every terrain-normalized experiment.

\begin{figure*}[!t]
\centering
\figorplaceholder{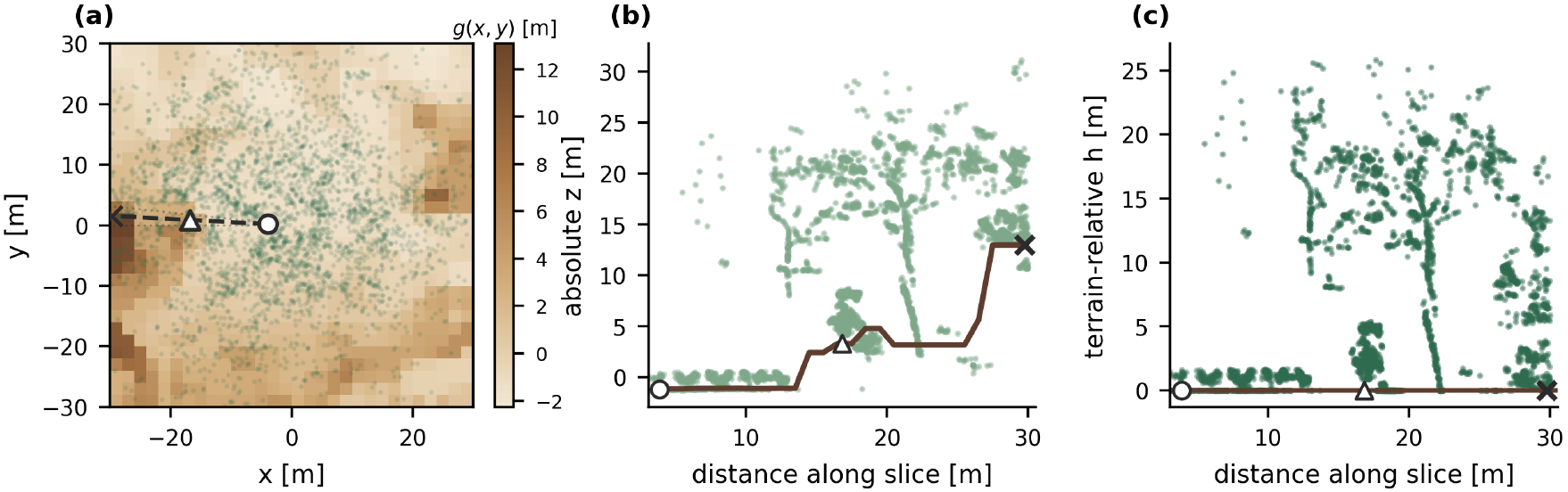}{0.86\textwidth}{1.12in}
\caption{Effect of terrain normalization on the vertical representation. The same local forest corridor is shown in absolute-height and terrain-relative coordinates. Removing low-frequency terrain relief aligns the local vertical reference while preserving vegetation-layer structure, which is subsequently encoded by TNVD.}
\label{fig:terrain}
\end{figure*}

Figure~\ref{fig:terrain} illustrates the role of this normalization: low-frequency terrain relief is removed from the vertical coordinate while local vegetation structure is preserved for the subsequent distribution encoding.

\subsection{Polar Vertical Distribution}
As in SC, the horizontal plane is divided into $N_R=20$ rings and $N_S=60$ sectors within an 80\,m radius. For a point assigned to cell $(r,s)$, terrain-relative height is quantized into $B=24$ bins over $[0,H]$, with $H=18$\,m and overflow accumulated in the final bin:
\begin{equation}
 b(h_i)=\min\!\left(B-1,\left\lfloor\frac{h_i}{H/B}\right\rfloor\right).
 \label{eq:height_bin}
\end{equation}
Let $n_{r,s,b}$ be the number of points in vertical bin $b$ and $N_{r,s}=\sum_{b=0}^{B-1}n_{r,s,b}$. We define the observation mask $O_{r,s}=\mathbb{I}[N_{r,s}>0]$ and
\begin{equation}
D_{r,s,b}=
\begin{cases}
\dfrac{n_{r,s,b}}{N_{r,s}}, & O_{r,s}=1,\\[4pt]
0, & O_{r,s}=0,
\end{cases}
\quad b=0,\ldots,B-1.
\label{eq:distribution}
\end{equation}
Thus $D_{r,s,:}$ is a categorical distribution only for an observed cell; an unobserved cell is represented by the zero vector and excluded from distribution-to-distribution comparison. TNVD preserves the relative probability mass across vertical layers while normalizing away the absolute return count. Section~\ref{sec:ablation} verifies that the probability mass contains information beyond binary vertical occupancy.

\subsection{Compact Yaw-Invariant Support Key}
For coarse retrieval, we summarize whether each vertical bin is present across the sectors of a ring:
\begin{equation}
 u_{r,b}=\frac{1}{N_S}\sum_{s=0}^{N_S-1}\mathbb{I}\!\left[n_{r,s,b}>0\right].
 \label{eq:support_key}
\end{equation}
The indicator is deliberately binary: a sector contributes one whenever at least one point occupies bin $b$, independent of the number of returns. Vectorizing $u_{r,b}$ produces a $N_RB=480$ dimensional key. Because the sector axis is aggregated, the key is invariant to circular shifts and, more strongly, arbitrary sector permutations. We accept this stronger invariance only at coarse retrieval; the full TNVD tensor retains sector ordering for candidate disambiguation. A KD-tree with Euclidean key distance retrieves $K=25$ candidates.

\subsection{Circular Jensen--Shannon Matching}
For two observed cell distributions $P,Q\in\Delta^{B-1}$, let $M=(P+Q)/2$. Using the natural logarithm,
\begin{equation}
\operatorname{JSD}(P,Q)
=\frac{1}{2}\operatorname{KL}(P\|M)
+\frac{1}{2}\operatorname{KL}(Q\|M),
\label{eq:jsd}
\end{equation}
where $\operatorname{KL}(P\|M)=\sum_{b:P_b>0}P_b\ln(P_b/M_b)$. Hence $0\leq\operatorname{JSD}(P,Q)\leq\ln2$, and the normalized cell similarity is
\begin{equation}
 c(P,Q)=1-\frac{\operatorname{JSD}(P,Q)}{\ln2}\in[0,1].
 \label{eq:cell_sim}
\end{equation}
For sector shift $\tau\in\{0,\ldots,N_S-1\}$, the database tensor is circularly shifted to the right,
\begin{equation}
\begin{aligned}
D^{d,(\tau)}_{r,s,b}
&=D^d_{r,(s-\tau)\bmod N_S,b},\\
O^{d,(\tau)}_{r,s}
&=O^d_{r,(s-\tau)\bmod N_S}.
\end{aligned}
\label{eq:db_shift}
\end{equation}
Define the active union
\begin{equation}
\mathcal{A}_{\tau}
=\{(r,s): O^q_{r,s}=1\ \lor\ O^{d,(\tau)}_{r,s}=1\}.
\label{eq:active_union}
\end{equation}
The active-cell contribution is
\begin{equation}
\gamma^{(\tau)}_{r,s}=
\begin{cases}
c\!\left(D^q_{r,s,:},D^{d,(\tau)}_{r,s,:}\right),
& O^q_{r,s}=O^{d,(\tau)}_{r,s}=1,\\[3pt]
0, & O^q_{r,s}\neq O^{d,(\tau)}_{r,s}.
\end{cases}
\label{eq:masked_cell_sim}
\end{equation}
Both-empty cells are absent from $\mathcal{A}_{\tau}$; one-sided cells enlarge the union but contribute zero. The shift score is
\begin{equation}
S(q,d;\tau)=\frac{1}{|\mathcal{A}_{\tau}|}
\sum_{(r,s)\in\mathcal{A}_{\tau}}\gamma^{(\tau)}_{r,s}.
\label{eq:shift_score}
\end{equation}
Finally,
\begin{subequations}\label{eq:best_alignment}
\begin{equation}
\hat{\tau}=\underset{0\leq\tau<N_S}{\arg\max}\,S(q,d;\tau),\qquad
S^*(q,d)=S(q,d;\hat{\tau}).
\label{eq:best_shift}
\end{equation}
\begin{equation}
\hat{\psi}_{q\rightarrow d}
=-\hat{\tau}\frac{2\pi}{N_S}\pmod{2\pi}.
\label{eq:yaw_alignment}
\end{equation}
\end{subequations}
Here $\hat{\psi}_{q\rightarrow d}$ denotes the discrete query-to-database circular alignment associated with the best descriptor match under the stated sector-index convention. With $N_S=60$, one sector corresponds to $6^{\circ}$. The candidate with the largest $S^*$ is returned as the recognized place.

\section{Experimental Evaluation}

\begin{table*}[!t]
\centering
\caption{Inter-session place recognition performance on Wild-Places under the standardized TreeLoc++ Table-5 protocol. Each site contains 12 directed query--database pairs. For each metric, we report the pair mean and stability ratio $S$ from Eq.~\eqref{eq:stability}. Published baselines are from TreeLoc++~\cite{jung2026treelocpp}. Best values are bold.}
\label{tab:wildplaces_main}
\vspace{-1mm}
\resizebox{\textwidth}{!}{%
\begin{tabular}{@{}ll|cc|cc|cc|cc||cc|cc|cc|cc@{}}
\toprule
\multirow{3}{*}{Category} &
\multirow{3}{*}{Method} &
\multicolumn{8}{c||}{Venman01--04 (12 query--database pairs)} &
\multicolumn{8}{c}{Karawatha01--04 (12 query--database pairs)} \\
\cmidrule(lr){3-10}\cmidrule(lr){11-18}
& &
\multicolumn{2}{c|}{R@1} &
\multicolumn{2}{c|}{MR} &
\multicolumn{2}{c|}{MF1} &
\multicolumn{2}{c||}{AUC} &
\multicolumn{2}{c|}{R@1} &
\multicolumn{2}{c|}{MR} &
\multicolumn{2}{c|}{MF1} &
\multicolumn{2}{c}{AUC} \\
& &
Mean & $S$ & Mean & $S$ & Mean & $S$ & Mean & $S$ &
Mean & $S$ & Mean & $S$ & Mean & $S$ & Mean & $S$ \\
\midrule
\multirow{6}{*}{Learning}
& TransLoc3D
& .553 & .464 & .008 & -.048 & .707 & .678 & .604 & .511
& .711 & 1.316 & .013 & -.103 & .831 & 1.561 & .756 & 1.268 \\
& LoGG3D-Net
& .660 & \textbf{1.242} & .003 & -.282 & .813 & \textit{1.382} & .732 & .872
& \textbf{.897} & \textbf{1.848} & .002 & -.278 & \textit{.948} & \textit{2.299} & .921 & \textit{1.638} \\
& MinkLoc3Dv2
& .755 & .782 & .033 & .161 & .857 & 1.030 & .818 & .881
& .849 & 1.607 & .062 & .150 & .919 & 1.889 & .911 & 1.592 \\
& BEVPlace++
& .735 & .701 & .091 & -.025 & .840 & .920 & .841 & .842
& .767 & .934 & .016 & -.193 & .868 & 1.179 & .767 & .822 \\
& ForestLPR
& .718 & .695 & .023 & .006 & .842 & 1.006 & .839 & 1.010
& .790 & 1.136 & .072 & .024 & .885 & 1.366 & .885 & 1.205 \\
& HOTFormerLoc
& \textbf{.899} & 1.173 & \textbf{.155} & -.057 & \textbf{.946} & 1.448 & \textbf{.938} & \textbf{1.350}
& .885 & 1.341 & \textbf{.088} & .026 & \textbf{.941} & 1.628 & \textbf{.937} & 1.471 \\
\midrule
\multirow{7}{*}{Handcrafted}
& SC++
& .245 & .429 & .056 & .129 & .425 & .631 & .396 & .430
& .370 & .632 & .117 & .146 & .560 & .817 & .564 & .704 \\
& SOLiD
& .150 & .199 & .014 & -.263 & .288 & .321 & .241 & .093
& .158 & .357 & .000 & .000 & .300 & .501 & .211 & .318 \\
& RING++
& .322 & .448 & .057 & -.004 & .503 & .624 & .500 & .458
& .341 & .713 & .030 & -.143 & .509 & .848 & .402 & .439 \\
& BTC
& .185 & .543 & .006 & -.337 & .331 & .594 & .216 & .338
& .068 & .169 & .005 & -.193 & .221 & .150 & .128 & .024 \\
& MapClosure
& .478 & .338 & .054 & -.008 & .618 & .459 & .465 & .279
& .484 & .420 & .036 & -.121 & .641 & .571 & .479 & .347 \\
& TreeLoc
& .766 & .940 & .046 & -.019 & .890 & 1.202 & .853 & .954
& .486 & .938 & .003 & .785 & .858 & 1.747 & .838 & 1.577 \\
& TreeLoc++
& .896 & 1.207 & \textbf{.970} & \textbf{1.469} & \textbf{.993} & 2.303 & \textbf{.998} & 3.281
& .728 & 1.266 & \textbf{.849} & \textbf{1.008} & \textbf{.974} & 2.149 & \textbf{.995} & 2.801 \\
\midrule
\multicolumn{2}{l|}{\textbf{TNVD-SC (ours)}}
& \textbf{.922} & \textbf{2.822} & .364 & .329 & .960 & \textbf{3.473} & .973 & \textbf{3.999}
& \textbf{.914} & \textbf{3.329} & .236 & .395 & .956 & \textbf{4.012} & .970 & \textbf{4.094} \\
\bottomrule
\end{tabular}%
}
\vspace{-1mm}
\begin{flushleft}
\footnotesize
\hspace{0.2em}\textit{R@1}: Recall@1; \textit{MR}: maximum recall at 100\% precision; \textit{MF1}: maximum F1 score; \textit{AUC}: area under the precision--recall curve; $S$: stability ratio.
\end{flushleft}
\vspace{-2mm}
\end{table*}

\subsection{Datasets and Protocols}
One \emph{place} is one input scan and one global descriptor for our SC and TNVD-SC rows; neither uses temporal accumulation. \textbf{Wild-Places:} we follow the TreeLoc++ inter-session protocol~\cite{jung2026treelocpp} on Venman01--04 and Karawatha01--04 (12 directed pairs/site), with the predefined test region, the full paired session as database, a 5\,m horizontal positive radius, and pair-wise means. \textbf{NCLT}~\cite{carlevaris2016nclt}: we reproduce the released SeqOT/CVTNet full-database evaluator~\cite{ma2023seqot,ma2023cvtnet}; 2012-01-08 (28,127 scans) is the database, four long-term sessions are queried every five scans, and positives use a 15\,m pose threshold. \textbf{Oxford Forest:} Evo, Forest of Dean, Stein am Rhein, and Wytham use the released TreeLoc++ Oxford data with a 5\,m positive radius and 50-frame exclusion. TreeLoc++ is re-run under the same evaluation rules and evaluated-query cohorts (1361/451/491/98). The frozen Wild-Places configuration \TNVDConfig{} is reused unchanged on NCLT and Oxford.

\subsection{Wild-Places: TreeLoc++ Standardized Comparison}
We follow the exact TreeLoc++ Table-5 reporting protocol~\cite{jung2026treelocpp}. For each site, the four sessions form 12 directed inter-session query--database pairs. Each metric is reported by its pair mean $\mu$ and the TreeLoc++ stability ratio
\begin{equation}
S=\log\left(\frac{\mu}{\sigma+\epsilon}\right),
\label{eq:stability}
\end{equation}
where $\sigma$ is the population standard deviation over the 12 directed pairs and $\epsilon=10^{-12}$ is the numerical stabilizer used by the Table-5 metric computation. For \ours{} R@1, the corresponding pair-wise standard deviations are approximately 0.0548 and 0.0327 on Venman and Karawatha, respectively, so $\epsilon$ is negligible relative to the measured variation. Table~\ref{tab:wildplaces_main} shows that \ours{} has the highest R@1 mean and stability on both sites, while TreeLoc++ remains stronger in mean MR, MF1, and AUC; we therefore claim top-1 accuracy/consistency rather than all-metric dominance. Among 12,350 top-1 true positives with query and database pose orientations available, the discrete circular alignment has a median absolute yaw error of 2.33$^\circ$ and a 90th percentile of 8.29$^\circ$ relative to the ground-truth query-to-database heading, supporting the geometric interpretation of the maximizing shift. The MR gap is mainly confidence separability: TreeLoc++ geometry-overlap confidence yields correctness ROC-AUC 0.998/0.990 (Venman/Karawatha), versus 0.790/0.768 for TNVD-SC similarity, motivating selective structural verification. Matched Filtering~\cite{joseph2025matched} and HOTFLoc++~\cite{griffiths2026hotflocpp} use different 3\,m Wild-Places protocols and are not mixed into Table~\ref{tab:wildplaces_main}.

\begin{table}[!t]
\centering
\caption{NCLT four-session CVTNet evaluation (R@1, \%). Published baselines are from CVTNet~\cite{ma2023cvtnet}; our rows use the released full-database evaluator.}
\label{tab:nclt}
\scriptsize
\setlength{\tabcolsep}{1.5pt}
\begin{tabular}{@{}lccccc@{}}
\toprule
Method & 02-05 & 06-15 & 02-23 & 04-05 & Mean \\
\midrule
PointNetVLAD & 74.6 & 61.2 & 46.9 & 44.9 & 56.90 \\
MinkLoc3D & 80.2 & 63.0 & 50.7 & 48.2 & 60.53 \\
OverlapTransformer & 86.1 & 63.9 & 53.6 & 49.6 & 63.30 \\
SeqOT & 91.7 & 76.2 & 69.1 & 63.9 & 75.23 \\
CVTNet & 93.2 & 79.3 & 81.7 & 78.0 & 83.05 \\
\midrule
SC (ours) & 91.89 & 69.09 & 70.15 & 63.77 & 73.73 \\
\textbf{TNVD-SC} & \textbf{93.38} & \textbf{81.61} & \textbf{84.59} & \textbf{80.97} & \textbf{85.14} \\
\bottomrule
\end{tabular}
\end{table}
\subsection{NCLT Long-Term Generalization}
Table~\ref{tab:nclt} evaluates the unchanged Wild-Places configuration with the released SeqOT/CVTNet full-database protocol. TNVD-SC improves our full-yaw SC baseline on every date, raising mean R@1 from \NCLTSCMean\% to \NCLTTNVDMean\% (+\NCLTGain\,pp) and exceeding CVTNet's reported four-session mean by 2.09 points. This claim is restricted to the reproduced CVTNet/SeqOT setup because other NCLT studies vary sessions, context, modality, or thresholds.

\begin{table}[!t]
\centering
\caption{Oxford Forest intra-session R@1 (\%) under the common 5\,m / 50-frame protocol. TreeLoc++ is re-run on the same released sequences and evaluated-query cohorts.}
\label{tab:oxford}
\scriptsize
\setlength{\tabcolsep}{3.0pt}
\begin{tabular}{@{}lrrrr@{}}
\toprule
Sequence & SC & TNVD-SC & TreeLoc++ & $\Delta$ \\
\midrule
Evo            & 30.49 & 40.78 & \textbf{93.90} & +10.29 \\
Forest of Dean & 22.39 & 73.17 & \textbf{76.27} & +50.78 \\
Stein am Rhein & 35.44 & 45.42 & \textbf{93.69} & +9.98 \\
Wytham         & 8.16  & \textbf{68.37} & 39.80 & +60.21 \\
\midrule
Mean           & 24.12 & 56.94 & \textbf{75.92} & +32.82 \\
\bottomrule
\end{tabular}
\end{table}
\subsection{Oxford Forest Generalization}
Table~\ref{tab:oxford} adds a TreeLoc++ re-run to the frozen Oxford control under the same 5\,m / 50-frame evaluated-query cohorts. TNVD-SC improves SC on all four sequences (24.12\% to 56.94\% mean), while TreeLoc++ leads overall at 75.92\%; however, TNVD-SC is close on Forest of Dean (73.17\% vs. 76.27\%) and higher on Wytham (68.37\% vs. 39.80\%), indicating complementary site-wise behavior.

\begin{table}[!t]
\centering
\caption{Retrieval-key, similarity, and representation ablations on standardized Wild-Places (R@1, \%). The full-distribution cosine entry intentionally repeats the cosine-similarity row as the reference condition for the binary-representation ablation.}
\label{tab:key_similarity}
\scriptsize
\setlength{\tabcolsep}{2.2pt}
\begin{tabular}{@{}llccc@{}}
\toprule
Ablation & Setting & Venman & Karawatha & Mean \\
\midrule
\multirow{5}{*}{Retrieval key}
& Mean, 480-D & 89.34 & 89.59 & 89.47 \\
& \textbf{Support only, 480-D} & 92.24 & \textbf{91.41} & 91.83 \\
& Mean+Std, 960-D & 88.04 & 87.82 & 87.93 \\
& Mean+Support, 960-D & 92.31 & 91.40 & 91.86 \\
& Mean+Std+Support, 1440-D & \textbf{92.37} & 91.36 & \textbf{91.87} \\
\midrule
\multirow{3}{*}{Similarity}
& Cosine & \textbf{92.38} & 91.30 & \textbf{91.84} \\
& Bhattacharyya & 90.87 & 90.07 & 90.47 \\
& \textbf{Jensen--Shannon} & 92.24 & \textbf{91.41} & 91.83 \\
\midrule
\multirow{2}{*}{Representation}
& Binary + cosine & 82.72 & 86.76 & 84.74 \\
& Full dist. + cosine & 92.38 & 91.30 & 91.84 \\
\bottomrule
\end{tabular}
\end{table}

\begin{table}[!t]
\centering
\caption{Vertical discretization sensitivity on standardized Wild-Places (R@1, \%).}
\label{tab:bh}
\scriptsize
\setlength{\tabcolsep}{2.8pt}
\begin{tabular}{@{}lcccc@{}}
\toprule
Setting & Value & Key dim. & Venman & Karawatha / Mean \\
\midrule
\multirow{4}{*}{$B$ ($H=18$ m)}
& 6  & 120 & 81.66 & 81.87 / 81.77 \\
& 12 & 240 & 89.21 & 88.60 / 88.91 \\
& \textbf{24} & \textbf{480} & \textbf{92.24} & \textbf{91.41 / 91.83} \\
& 36 & 720 & 93.34 & 92.39 / 92.87 \\
\midrule
\multirow{4}{*}{$H$ ($B=24$)}
& 12 m & 480 & 92.32 & 89.90 / 91.11 \\
& \textbf{18 m} & \textbf{480} & \textbf{92.24} & \textbf{91.41 / 91.83} \\
& 24 m & 480 & 91.62 & 91.03 / 91.33 \\
& 30 m & 480 & 90.58 & 90.37 / 90.48 \\
\bottomrule
\end{tabular}
\end{table}
\subsection{Ablations and Design Choices}
\label{sec:ablation}
Table~\ref{tab:key_similarity} separates the requirements of coarse retrieval from those of fine matching. The 480-D support-only key reaches 91.83\% mean R@1, only 0.04 points below the 1440-D mean+std+support key, indicating that occupancy across vertical bins is already sufficient for candidate generation. In contrast, the ordered matcher benefits strongly from retaining probability mass: binarizing the full distribution lowers mean R@1 from 91.84\% to 84.74\%. Cosine and \JS{} matching are nearly tied in raw accuracy (91.84\% versus 91.83\%), so \JS{} is not selected because of an isolated accuracy advantage. We use it because the cell descriptor is a normalized distribution and because its divergence structure supports the exact hierarchical pruning described below.

Table~\ref{tab:bh} further shows that most of the discretization gain is obtained before $B=24$. Increasing $B$ from 6 to 12 and then 24 raises mean R@1 from 81.77\% to 88.91\% and 91.83\%, whereas $B=36$ adds only 1.04 points while increasing retrieval-key dimension by 50\%. We therefore retain $B=24$ as the accuracy--compactness knee. With $B$ fixed, performance is comparatively stable around $H=18$\,m and degrades as the range becomes too large, consistent with a fixed bin budget allocating less resolution to the vegetation layers that are most discriminative. These ablations freeze $B=24$, $H=18$\,m, support-only retrieval, and full-distribution \JS{} matching for all reported generalization experiments.

\subsection{Implementation Efficiency}
\label{sec:efficiency}
The TNVD tensor is reconstructed exactly from \texttt{uint16} bin counts, reducing persistent payload from 239,040 to 61,920 bytes/place (3.86$\times$). This representation is lossless with respect to the normalized TNVD probabilities because each probability is recovered from the stored per-bin counts and the corresponding cell total; no quantized probability approximation is introduced.

Exact hierarchical \JS{} search reduces matching cost without changing the objective. Adjacent vertical bins are first aggregated into a coarser distribution and used only to reject yaw shifts that cannot improve the current best score. Surviving shifts are always evaluated with the original $B=24$ distributions, so the hierarchy changes the order and amount of computation rather than the final similarity definition or yaw resolution. The exhaustive and hierarchical implementations agree on all 13,459 evaluation queries. Table~\ref{tab:efficiency} shows compute latency falling from 228.55 to 174.27\,ms (1.312$\times$) and mean DB-resident memory from 504.85 to 188.50\,MiB; a separate 4,000-query audit preserves every top-1 place and yaw decision. For context, conventional SC averages 18.73\,ms/query under the same Wild-Places $K=25$, 500-query-per-database clean benchmark protocol. The optimized TNVD-SC is therefore about 9.3$\times$ more expensive than SC but still processes 5.74 queries/s on the measured CPU, which supports periodic loop-closure or relocalization queries without requiring execution at every LiDAR frame.

\begin{table}[!t]
\centering
\caption{Implementation efficiency. SC timing is reported as a reference from the same $K=25$, 500-query clean benchmark; its DB-RSS value was not recorded with the final compact-memory campaign.}
\label{tab:efficiency}
\scriptsize
\setlength{\tabcolsep}{2.2pt}
\begin{tabular}{@{}lrrrr@{}}
\toprule
Implementation & Payload [kB] & DB [MiB] & Query [ms] & Rate [Hz] \\
\midrule
SC reference & \textbf{9.7} & -- & \textbf{18.73} & \textbf{53.4} \\
TNVD float exhaustive & 239.0 & 504.9 & 228.55 & 4.38 \\
TNVD count + exact hier. & 61.9 & 188.5 & 174.27 & 5.74 \\
\bottomrule
\end{tabular}
\vspace{-1.5mm}
\end{table}

\subsection{Residual Failure Analysis and Spatial Recovery}
\label{sec:qualitative}
On the strict Wild-Places cohort, 98.73\% of queries retain a positive in the top-25. Thus, after the support key has formed the candidate set, the dominant residual problem is no longer candidate omission: 85.1\% of the remaining failures are reranking errors. Moreover, 52.6\% of those failures select a false location more than 20\,m away, showing that the difficult cases are often genuine long-range aliases rather than small threshold-boundary errors. This separation is important because simply increasing $K$ cannot resolve a false candidate that already receives a higher fine-matching score than the correct place.

\begin{figure*}[!t]
\centering
\figorplaceholder{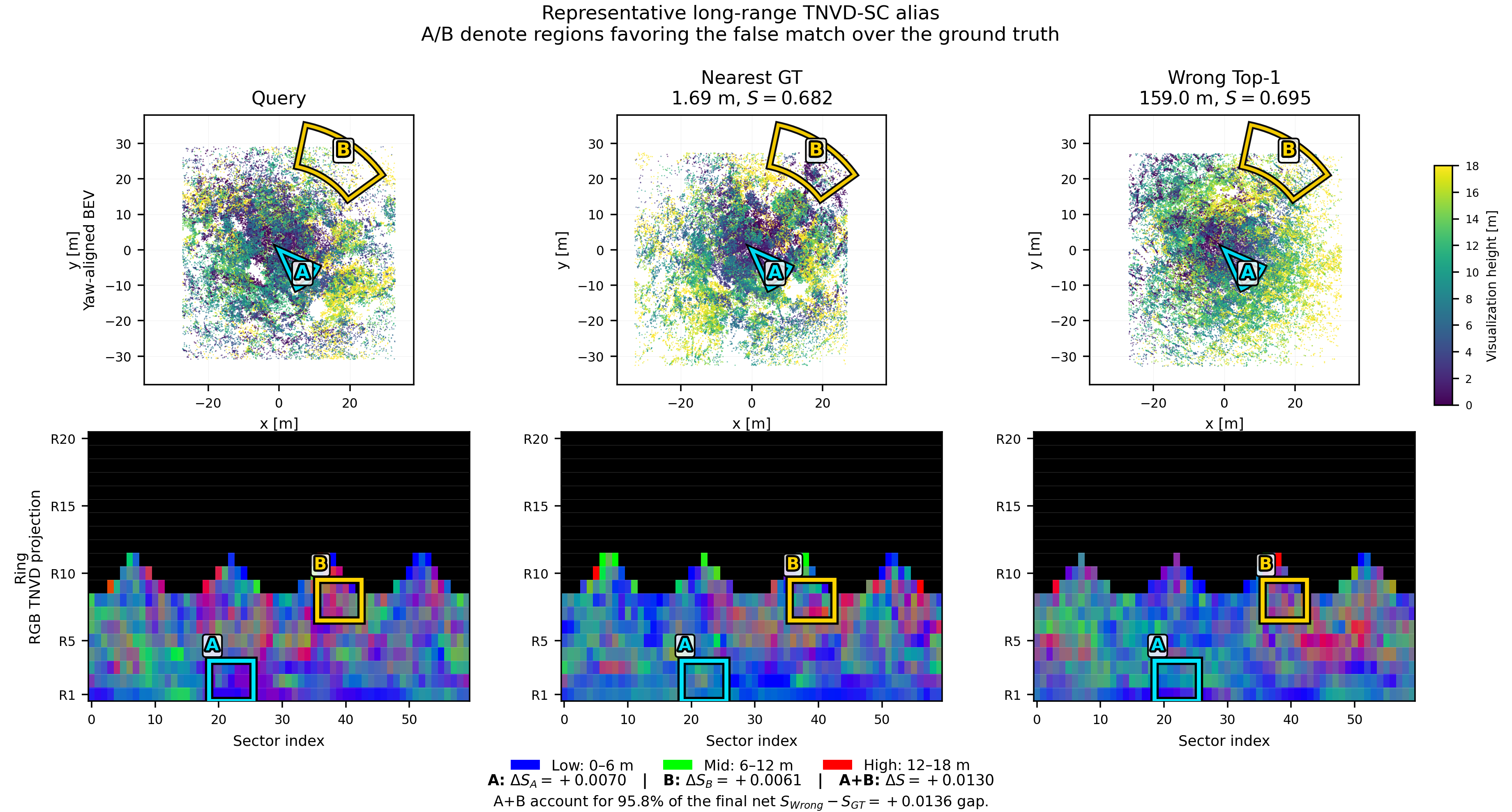}{0.94\textwidth}{1.60in}
\caption{Long-range TNVD-SC alias. A 159.0\,m false candidate outranks the 1.69\,m GT; regions A/B explain 95.8\% of the net similarity gap. The lower row collapses the 24-bin TNVD tensor into three height bands (0--6, 6--12, and 12--18\,m) for display only; matching uses all $B=24$ bins.}
\label{fig:residual_alias}
\vspace{-1.5mm}
\end{figure*}

Figure~\ref{fig:residual_alias} illustrates such a case. A candidate 159.0\,m from the query outranks the 1.69\,m ground-truth match because two spatial regions with similar terrain-relative vertical layouts contribute most of the net similarity advantage. Regions A/B account for 95.8\% of the score gap, indicating a localized structural alias rather than diffuse descriptor noise. TNVD-SC therefore removes many max-height ambiguities but does not make distinct forest locations uniquely identifiable when their internal vertical organization is itself similar.

\begin{figure*}[!t]
\centering
\figorplaceholder{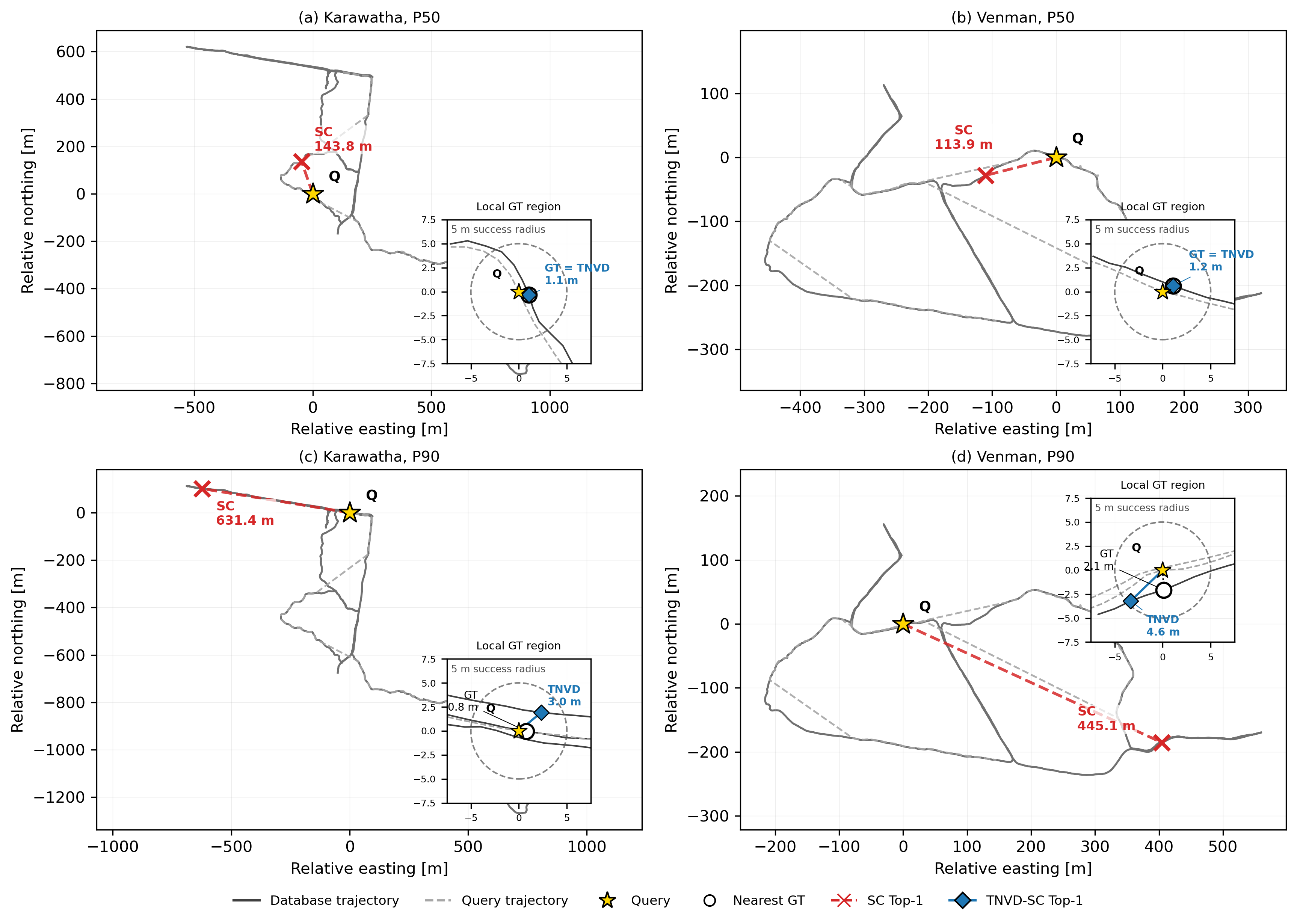}{0.71\textwidth}{1.95in}
\caption{SC failures recovered by TNVD-SC. P50/P90 examples use representative pairs nearest to each site's mean gain; insets show the 5\,m GT region.}
\label{fig:trajectory_rescue}
\vspace{-1.5mm}
\end{figure*}

The complementary behavior is shown in Fig.~\ref{fig:trajectory_rescue}. Four representative SC failures lie 113.9--631.4\,m from the query, yet TNVD-SC recovers matches within 1.1--4.6\,m. In these examples, distant places produce similar upper envelopes under scalar max-height pooling, whereas terrain-relative vertical distributions preserve differences in lower and mid-level vegetation occupancy. The qualitative cases therefore mirror the controlled ablations: the main gain comes from preserving within-cell vertical mass for reranking, not from widening the candidate search.

\section{Discussion}
The controlled experiments separate candidate recovery from fine discrimination. The support key nearly saturates Top-25 positive coverage, yet conventional SC matching still loses many retrieved positives; terrain normalization is likewise most effective when the downstream cell representation preserves vertical probability mass. The gain therefore follows the diagnosed fine-matching bottleneck rather than retrieval breadth or ground normalization alone.

Residual failures are mostly reranking ambiguities, and Fig.~\ref{fig:residual_alias} shows that similar vertical layouts can still outrank the correct place. TreeLoc++ remains stronger in mean Wild-Places MR/MF1/AUC and Oxford R@1, while TNVD-SC wins some Oxford sequences, supporting complementary representations. We leave confidence-aware structural routing outside the present method and use NCLT/Oxford to test the frozen TNVD representation directly.

First-bin saturation remains a limitation: the zero-clamped first bin carries 29.9\%/31.4\% mean occupied-cell probability mass on Venman/Karawatha. A matcher-only control that removes this bin while retaining the frozen 480-D retrieval key improves strict 24-pair mean R@1 by 2.43 points; because this refinement was evaluated only on Wild-Places, we retain the all-bin configuration for generalization experiments.

\section{Conclusion}
We presented \ours{}, a training-free SC-compatible system derived from a stage-wise diagnosis of forest aliasing. The results show that the dominant gain comes from preserving terrain-relative within-cell vertical structure for fine matching, while a compact support key is sufficient for retrieval. The frozen method reaches 91.83\% mean R@1 on standardized Wild-Places, transfers to NCLT and Oxford Forest, and supports exact count-coded hierarchical matching. The unchanged cross-domain evaluation further indicates that this separation is not tied to the Wild-Places tuning domain.

\vspace{-1mm}
\section*{Acknowledgment}
\vspace{-1mm}
OpenAI ChatGPT was used to assist with manuscript drafting and language refinement and to generate and iteratively refine Fig.~\ref{fig:architecture} from author-specified technical content; the authors reviewed and verified the final text and figure for technical accuracy.

\FloatBarrier

\providecommand{\BIBdecl}{}
\renewcommand{\BIBdecl}{\fontsize{7.3pt}{8.0pt}\selectfont}
% Generated by IEEEtran.bst, version: 1.14 (2015/08/26)

\end{document}